\documentclass[twoside]{article}
\usepackage{microtype}
\usepackage{graphicx}
\usepackage{comment}
\usepackage{subcaption}
\usepackage{booktabs} 
\usepackage{multirow}
\usepackage{algorithm}
\usepackage{algorithmic}
\usepackage[utf8]{inputenc} 
\usepackage[T1]{fontenc}    
\usepackage{url}            
\usepackage{booktabs}       
\usepackage{amsfonts}       
\usepackage{nicefrac}       
\usepackage{microtype}      
\usepackage[utf8]{inputenc}
\usepackage{float}
\usepackage[T1]{fontenc}    
\usepackage{hyperref}       
\usepackage{url}            
\usepackage{booktabs}       
\usepackage{amsfonts}       
\usepackage{nicefrac}       
\usepackage{microtype}      
\usepackage{graphicx}
\usepackage{wrapfig}
\usepackage{amsthm}
\usepackage{placeins}
\usepackage{hyperref}
\usepackage[english]{babel}
\usepackage{url}
\usepackage{mathtools}
\usepackage{amsmath}
\usepackage{amssymb}
\usepackage{wrapfig}
\usepackage{bm}
\usepackage{psfrag}
\usepackage{tikz}
\usepackage{epstopdf}
\usepackage{xcolor}
\usepackage{adjustbox}

\newcommand{\beq}[1][\vspace{0.3em}]{#1\begin{equation}}
\newcommand{\eeq}{\end{equation}}

\newcommand{\bit}{\vspace{0mm}\begin{itemize}}
\newcommand{\eit}{\vspace{0mm}\end{itemize}}
\newcommand{\ben}{\vspace{0mm}\begin{enumerate}}
\newcommand{\een}{\vspace{0mm}\end{enumerate}}
\newtheorem{theorem}{Theorem}

\newtheorem{corollary}[theorem]{Corollary}
\newtheorem{lemma}[theorem]{Lemma}

\newcommand{\Cv}[0]{{{\bf C}}}

\newcommand{\uv}[0]{{{\bf u}}}

\newcommand{\xv}[0]{{{\bf x}}}

\newcommand{\zv}[0]{{{\bf z}}}
\newcommand{\epsilonv}[0]{{{\bm{\epsilon}}}}
\newcommand{\lambdav}[0]{{{\bm{\lambda}}}}

\DeclareMathOperator*{\argmin}{arg\,min}

\newcommand{\bb}[1]{\mathbb{#1}}

\newcommand{\mc}[1]{\mathcal{#1}}

\usepackage[colorinlistoftodos]{todonotes}
\usepackage{soul}
\usepackage{array}
\newcolumntype{C}{>{\centering\arraybackslash}p{1em}}
\providecommand{\js}[1]{{\color{cyan} [JS: #1]}}
\providecommand{\s}[1]{{\color{blue} [SE: #1]}}

\providecommand{\rk}[1]{{\color{orange} [RK: #1]}}
\providecommand{\done}[1]{{}}
\providecommand{\x}[1]{{}} 

\providecommand{\tmp}[1]{{}}
\providecommand{\xx}[1]{{}} 

\newcommand{\iq}[0]{{{I_q}}}
\newcommand{\hq}[0]{{{H_q}}}

\newcommand{\pxizu}[0]{{{p_\theta(\xv|\zv, \uv)}}}
\newcommand{\qzixu}[0]{{{q_\phi(\zv|\xv, \uv)}}}
\newcommand{\qxizu}[0]{{{q_\phi(\xv|\zv, \uv)}}}
\newcommand{\quiz}[0]{{{q_\phi(\uv|\zv)}}}
\newcommand{\puiz}[0]{{{p_\psi(\uv|\zv)}}}

\newcommand{\qxiu}[0]{{{q(\xv|\uv)}}}
\newcommand{\qziu}[0]{{{q_\phi(\zv|\uv)}}}

\newcommand{\pu}[0]{{{p(\uv)}}}
\newcommand{\pz}[0]{{{p(\zv)}}}

\newcommand{\qu}[0]{{{q(\uv)}}}

\newcommand{\qz}[0]{{{q_\phi(\zv)}}}
\newcommand{\qxu}[0]{{{q(\xv, \uv)}}}
\newcommand{\qzu}[0]{{{q_\phi(\zv, \uv)}}}

\newcommand{\xui}[0]{{{(\xv_i, \uv_i)}}}
\newcommand{\xuyi}[0]{{{(\xv_i, \uv_i, y_i)}}}

\newcommand{\qxziu}[0]{{{q_\phi(\xv, \zv | \uv)}}}

\newcommand{\qxzu}[0]{{{q_\phi(\xv, \zv, \uv)}}}

\newcommand{\quizy}[0]{{{q_\phi(\uv|\zv, y)}}}

\newcommand{\KL}{D_{\mathrm{KL}}}

\newcommand{\methodlong}{{Mutual Information-based Fair Representations}}
\newcommand{\method}{{MIFR}}
\usepackage[accepted]{aistats2019}
\usepackage[round]{natbib}

\renewcommand{\s}[1]{{\color{red} [SE: #1]}}

\begin{document}

%

%
\runningauthor{ Jiaming Song$^\star$, Pratyusha Kalluri$^\star$, Aditya Grover, Shengjia Zhao, Stefano Ermon }

\twocolumn[

\aistatstitle{Learning Controllable Fair Representations}

\aistatsauthor{ Jiaming Song$^\star$ \And Pratyusha Kalluri$^\star$ \And Aditya Grover \And Shengjia Zhao \And Stefano Ermon }

\aistatsaddress{ Computer Science Department, Stanford University } ]

\begin{abstract}

Learning data representations that are transferable and are fair with respect to certain protected attributes is crucial to reducing unfair decisions while preserving the utility of the data. We propose an information-theoretically motivated objective for learning maximally expressive representations subject to fairness constraints. We demonstrate that a range of existing approaches optimize approximations to the Lagrangian dual of our objective. In contrast to these existing approaches, our objective allows the user to control the fairness of the representations by specifying limits on unfairness. Exploiting duality, we introduce a method that optimizes the model parameters as well as the expressiveness-fairness trade-off. Empirical evidence suggests that our proposed method can balance the trade-off between multiple notions of fairness and achieves higher expressiveness at a lower computational cost.
\end{abstract}

\section{INTRODUCTION}

Statistical learning systems are increasingly being used to assess individuals, influencing consequential decisions such as bank loans, college admissions, and criminal sentences.
This yields a growing demand for systems guaranteed to output decisions that are fair with respect to sensitive attributes such as gender, race, and disability. 

In the typical classification and regression settings with fairness and privacy constraints, one is concerned about performing a single, specific task.
However, situations arise where a data owner needs to release data to downstream users without prior knowledge of the tasks that will be performed~\citep{madras2018learning}. In such cases, it is crucial to find representations of the data that can be used on a wide variety of tasks while preserving fairness~\citep{calmon2017optimized}. 


This gives rise to two desiderata. On the one hand, the representations need to be \textit{expressive}, so that they can be used effectively for as many tasks as possible. On the other hand, the representations also need to satisfy certain \textit{fairness} constraints to protect sensitive attributes. 
Further, many notions of fairness are possible,
and it may not be possible to simultaneously satisfy all of them~\citep{kleinberg2016inherent,chouldechova2017fair}. Therefore, the ability to effectively trade off multiple notions of fairness is crucial to fair representation learning.


To this end, we present an information theoretically motivated constrained optimization framework (Section~\ref{sec:mi}). The goal is to maximize the expressiveness of representations while satisfying certain fairness constraints.
We represent expressiveness as well as three dominant notions of fairness (demographic parity~\citep{zemel2013learning}, equalized odds, equalized opportunity~\citep{hardt2016equality}) 
in terms of mutual information, obtain tractable upper/lower bounds of these mutual information objectives, and connect them with existing objectives such as maximum likelihood, adversarial training~\citep{goodfellow2014generative}, and variational autoencoders~\citep{kingma2013auto,rezende2015variational}.


As we demonstrate in Section~\ref{sec:framework}, this serves as a unifying framework for existing work~\citep{zemel2013learning,louizos2015the,edwards2015censoring,madras2018learning} on learning fair representations. A range of existing approaches to learning fair representations, which do not draw connections to information theory, optimize an approximation of the Lagrangian dual of our objective with fixed values of the Lagrange multipliers. These thus require the user to obtain different representations for different notions of fairness as in~\citet{madras2018learning}.


Instead, we consider a dual optimization approach (Section~\ref{sec:dual}), in which we optimize the model as well as the Lagrange multipliers during training~\citep{zhao2018a}, thereby also learning the trade-off between expressiveness and fairness. We further show that our proposed framework is strongly convex in distribution space.

Our work is the first
to provide direct user control over the fairness of representations through fairness constraints that are interpretable by non-expert users.
Empirical results in Section~\ref{sec:experiments} demonstrate that our notions of expressiveness and fairness based on mutual information align well with existing definitions, our method encourages representations that satisfy the fairness constraints while being more expressive, and that our method is able to balance the trade-off between multiple notions of fairness with a single representation and a significantly lower computational cost.
 \section{AN INFORMATION-THEORETIC OBJECTIVE FOR CONTROLLABLE FAIR REPRESENTATIONS}
 \label{sec:mi}

We are given a dataset $\mc{D}_u = \{\xui\}_{i=1}^{M}$ containing pairs of observations $\xv \in \mc{X}$ and sensitive attributes $\uv \in \mc{U}$. We assume the dataset is sampled i.i.d. from an unknown data distribution $\qxu$. 
Our goal is to transform each data point $(\xv, \uv)$ into a new representation $\zv \in \mc{Z}$ that 
is (1) \emph{transferable}, i.e., it can be used in place of $(\xv, \uv)$ by multiple unknown vendors on a variety of downstream tasks, and (2) \emph{fair}, i.e., the sensitive attributes $\uv$ are protected.
For conciseness, we focus on the \textit{demographic parity} notion of fairness \citep{calders2009building,zliobaite2015on,zafar2015learning}, which requires the decisions made by a classifier over $\zv$ to be independent of the sensitive attributes $\uv$. We discuss in Appendix~\ref{app:eodds-eopp} how our approach can be extended to control other notions of fairness simultaneously, such as the \textit{equalized odds} and \textit{equalized opportunity} notions of fairness~\citep{hardt2016equality}.

We assume the representations $\zv \in \mc{Z}$ of $(\xv, \uv)$ are obtained by sampling from a conditional probability distribution $\qzixu$ parameterized by $\phi \in \Phi$. The joint distribution of $(\xv, \zv, \uv)$ is then given by $\qxzu = \qxu \qzixu$.
We formally express our desiderata for learning a controllable fair representation $\zv$ through the concept of mutual information:
%
\begin{enumerate}
\item \textbf{Fairness}: \x{The representation} $\zv$ should have low mutual information with the sensitive attributes $\uv$.
\item \textbf{Expressiveness}: \x{The representation} $\zv$ should have high mutual information with the observations $\xv$, conditioned on $\uv$ 
(in expectation over possible values of $\uv$).
\end{enumerate}
The first condition encourages $\zv$ to be independent of $\uv$; if this is indeed the case, the downstream vendor cannot learn a classifier over the representations $\zv$ that discriminates based on $\uv$. Intuitively, the mutual information $\iq(\zv, \uv)$ is related to the optimal predictor of $\uv$ given $\zv$. If $\iq(\zv, \uv)$ is zero, then no such predictor can perform better than chance; if $\iq(\zv, \uv)$ is large, vendors in downstream tasks could utilize $\zv$ to predict the sensitive attributes $\uv$  and make unfair decisions.

The second condition encourages $\zv$ to contain as much information as possible from $\xv$ conditioned on the knowledge of $\uv$. By conditioning on $\uv$, we ensure we do not encourage information in $\xv$ that is correlated with $\uv$ to leak into $\zv$. The two desiderata allow $\zv$ to encode non-sensitive information from $\xv$ (expressiveness) while excluding information in $\uv$ (fairness).




\tmp{\s{does it ever make sense to have stochastic mapping to features?}}

Our goal is to choose parameters $\phi \in \Phi$ for $\qzixu$ that meet both these criteria\footnote{Simply ignoring $\uv$ as an input is insufficient, as $\xv$ may still contain information about $\uv$.}. Because we wish to ensure our representations satisfy fairness constraints even at the cost of using less expressive $\zv$, we synthesize the two desiderata into the following constrained optimization problem:
\begin{gather} 
\max_{\phi \in \Phi} \iq(\xv; \zv | \uv) \qquad \qquad \text{s.t. } \iq(\zv; \uv) < \epsilon \label{eq:constraint-objective}
\end{gather}
where $\iq(\xv; \zv | \uv)$ denotes the mutual information of $\xv$ and $\zv$ conditioned on $\uv$,
$\iq(\zv; \uv)$ denotes mutual information between $\zv$ and $\uv$, and the hyperparameter $\epsilon > 0$ controls the maximum amount of mutual information allowed between $\zv$ and $\uv$. The motivation of our ``hard'' constraint on $I_q(\zv; \uv)$ -- as opposed to a ``soft'' regularization term -- 
is that even at the cost of learning less expressive $\zv$ and losing some predictive power, we view as important ensuring that our representations are fair to the extent dictated by $\epsilon$. 

Both mutual information terms in Equation~\ref{eq:constraint-objective} are difficult to compute and optimize. 
In particular, the optimization objective in Equation~\ref{eq:constraint-objective} can be expressed as the following expectation:
\begin{align*}
&\iq(\xv; \zv | \uv) \\
= & \ \bb{E}_{\qxzu}[\log \qxziu - \log \qxiu - \log \qziu]    
\end{align*}
while the constraint on $\iq(\zv; \uv)$ involves the following expectation:
\begin{align*}
\iq(\zv; \uv) =  \ \bb{E}_{\qzu}[\log \qziu - \log \qz]    
\end{align*}
%
Even though $\qzixu$ is known analytically and assumed to be easy to evaluate, both mutual information terms are difficult to estimate and optimize. 

To offset the challenge in estimating mutual information, we introduce upper and lower bounds with tractable Monte Carlo gradient estimates. We introduce the following lemmas, with the proofs provided in Appendix~\ref{sec:proofs}. We note that similar bounds have been proposed in \citet{alemi2016deep,alemi2017fixing,zhao2018a,grover2019uae}. 


\subsection{Tractable Lower Bound for $\iq(\xv; \zv | \uv)$}

We begin with a (variational) \emph{lower} bound on the objective  function $\iq(\xv; \zv | \uv)$ related to expressiveness which we would like to \emph{maximize} in Equation~\ref{eq:constraint-objective}. 
\begin{lemma} \label{thm:reconstruction}
For any conditional distribution $p_\theta(\xv|\zv, \uv)$ (parametrized by $\theta$)
\begin{align*}
\iq(\xv; \zv | \uv) &= \bb{E}_{\qxzu}[\log p_\theta(\xv|\zv, \uv)] + \hq(\xv | \uv)\\
\nonumber
& \quad + \bb{E}_{\qzu}\KL(\qxizu \Vert p_\theta(\xv|\zv, \uv))
\end{align*}
where $\hq(\xv|\uv)$ is the entropy of $\xv$ conditioned on $\uv$, and $\KL$ denotes KL-divergence. 
\end{lemma}

Since entropy and KL divergence are non-negative, the above lemma implies the following lower bound:
\begin{align}
\label{eq:mi:lb}
\iq(\xv; \zv | \uv) &\geq \bb{E}_{\qxzu}[\log p_\theta(\xv|\zv, \uv)]:= \mc{L}_r.
\end{align}

\subsection{Tractable Upper Bound for $\iq(\zv; \uv)$}
Next, we provide an \emph{upper} bound for the constraint term $\iq(\zv; \uv)$ that specifies the limit on unfairness. In order to satisfy this fairness constraint, we wish to implicitly \emph{minimize} this term.
\begin{lemma} \label{thm:elbo}
For any distribution $\pz$, we have:
\begin{align} \label{eq:mi:12} 
 & \iq(\zv; \uv) \leq \iq(\zv; \xv, \uv) \\
= \ & \bb{E}_{\qxu}\KL(\qzixu \Vert \pz) - \KL(\qz \Vert \pz). \nonumber
\end{align}
\end{lemma}

Again, using the non-negativity of KL divergence, we obtain the following upper bound:
\begin{align} 
\iq(\zv; \uv)  \leq \bb{E}_{\qxu}\KL(\qzixu \Vert \pz):=C_1. \label{eq:mi:2}
\end{align}





In summary, Equation~\ref{eq:mi:lb} and Equation~\ref{eq:mi:2} imply that we can compute tractable Monte Carlo estimates for the lower and upper bounds to $\iq(\xv; \zv | \uv)$ and $\iq(\zv; \uv)$ respectively, as long as the variational distributions $p(\xv|\zv, \uv)$ and $\pz$ can be evaluated tractably, e.g., Bernoulli and Gaussian distributions. Note that the distribution $\qzixu$ is assumed to be tractable.


\subsection{A Tighter Upper Bound to $\iq(\zv, \uv)$ via Adversarial Training}


It would be tempting to use $C_1$, the tractable upper bound from Equation~\ref{eq:mi:2}, as a replacement for $\iq(\zv, \uv)$ in the constraint of Equation~\ref{eq:constraint-objective}. However, note from Equation~\ref{eq:mi:12} that $C_1$ is \emph{also} an upper bound to $\iq(\xv, \zv|\uv)$, which is the objective function (expressiveness) we would like to maximize in Equation~\ref{eq:constraint-objective}. If this was constrained too tightly, we would constrain the expressiveness of our learned representations. Therefore, we introduce a tighter bound via the following lemma.


\begin{lemma} \label{thm:adversarial}
For any distribution $\pu$, we have:
\begin{align}\label{eq:mi:ub:lemma3}
\iq(\zv; \uv) = & \ \bb{E}_{\qz}\KL(\quiz \Vert \pu) - \KL(\qu \Vert \pu). 
\end{align}
\end{lemma}

Using the non-negativity of KL divergence as before, we obtain the following upper bound on $\iq(\zv; \uv)$:
\begin{align}\label{eq:mi_c2_hat}
    \iq(\zv; \uv) \leq \bb{E}_{\qz}\KL(\quiz \Vert \pu) := \hat{C}_2.
\end{align}

As $\uv$ is typically low-dimensional (e.g., a binary variable, as in~\citet{hardt2016equality,zemel2013learning}), we can choose $\pu$ in Equation~\ref{eq:mi:ub:lemma3} to be a kernel density estimate based on the dataset $\mc{D}$. 
By making $\KL(\qu \Vert \pu)$ as small as possible, our upper bound $\hat{C}_2$ gets closer to $\iq(\zv, \uv)$.

While $\hat{C}_2$ is a valid upper bound to $\iq(\zv; \uv)$, the term $\quiz$ appearing in $\hat{C}_2$ is intractable to evaluate, requiring an integration over $\xv$. Our solution is to approximate $\quiz$ with a parametrized model $\puiz$ with parameters $\psi \in \Psi$ obtained via the following objective:
\begin{equation}
     \min_\psi \bb{E}_{\qz}\KL(\quiz \Vert \puiz). 
\end{equation}
Note that the above objective corresponds to maximum likelihood prediction with inputs $\zv$ and labels $\uv$ using $\puiz$. In contrast to $\quiz$, the distribution $\puiz$ is tractable and implies the following lower bound to $\hat{C}_2$: 
\begin{align*}
    & \bb{E}_{\qzu}[\log \puiz - \log \pu] \\
    = & \ \bb{E}_{\qz}[\KL(\quiz \Vert \pu) - \KL(\quiz \Vert \puiz)] \\
    \leq & \ \bb{E}_{\qz}\KL(\quiz \Vert \pu) = \hat{C}_2.
\end{align*}
It follows that we can approximate $\iq(\zv; \uv)$ through the following adversarial training objective:
\begin{equation}
    \min_\phi \max_\psi \bb{E}_{\qzu}[\log \puiz - \log \pu]
\end{equation}
Here, the goal of the adversary $p_\psi$ is to minimize the difference between the tractable approximation given by $\bb{E}_{\qzu}[\log \puiz - \log \pu]$ and the intractable true upper bound $\hat{C}_2$.
We summarize this observation in the following result:
\begin{corollary} \label{thm:adversarial-gap}
If $\KL(\quiz \Vert \puiz) \leq \ell$, then 
\begin{align*}
    \iq(\zv; \uv) & \leq \bb{E}_{\qzu}[\log \puiz - \log \pu] \\
    & \quad - \KL(\qu \Vert \pu) + \ell 
\end{align*}
for any distribution $p(\uv)$. 
\end{corollary}
It immediately follows that when $\ell \to 0$, i.e., the adversary approaches global optimality, we obtain the true upper bound. For any other finite value of $\ell$, we have:
\begin{align}
    \iq(\zv; \uv) &\leq \bb{E}_{\qzu}[\log \puiz - \log \pu]  + \ell\nonumber\\
    &:= C_2 + \ell.
\end{align}

\subsection{A practical objective for controllable fair representations}
Recall that our goal is to find tractable estimates to the mutual information terms in Equation~\ref{eq:constraint-objective} to make the objective and constraints tractable. In the previous sections, we have derived a lower bound for $\iq(\xv, \uv|\zv)$ (which we want to maximize) and upper bounds for $\iq(\uv, \zv)$ (which we want to implicitly minimize to satisfy the constraint). Therefore, by applying these results to the optimization problem in Equation~\ref{eq:constraint-objective}, we obtain the following constrained optimization problem:
\begin{align}
 \min_{\theta, \phi} \max_{\psi \in \Psi} \quad  &  \mc{L}_r = - \bb{E}_{\qxzu}[\log \pxizu] \label{eq:tractable-objective} \\
\text{s.t.} \quad & C_1 = \bb{E}_{\qxu}\KL(\qzixu \Vert \pz) < \epsilon_1 \nonumber \\
    & C_2 = \bb{E}_{\qzu}[\log \puiz - \log \pu] < \epsilon_2 \nonumber
\end{align}
where $\mc{L}_r$, $C_1$, and $C_2$ are introduced in
 Equations~\ref{eq:mi:lb},~\ref{eq:mi:2} and \ref{eq:mi_c2_hat} respectively.
Both $C_1$ and $C_2$ provide a way to limit $\iq(\zv, \uv)$.  
$C_1$ is guaranteed to be an upper bound to $\iq(\zv; \uv)$ but also upper-bounds $\iq(\xv; \zv | \uv)$ (which we would like to maximize), so it is more suitable when we value true guarantees on fairness over expressiveness. 
$C_2$ may more accurately approximate $\iq(\zv; \uv)$ but is guaranteed to be an upper bound only in the case of an optimal adversary. Hence, it is more suited for scenarios where the user is satisfied with guarantees on fairness in the limit of adversarial training, and we wish to learn more expressive representations. Depending on the underlying application, the user can effectively remove either of the constraints $C_1$ or $C_2$ (or even both) by setting the corresponding $\epsilon$ to infinity. 
\section{A UNIFYING FRAMEWORK FOR RELATED WORK}
\label{sec:framework}

Multiple methods for learning fair representations have been proposed in the literature. \citet{zemel2013learning} propose a method for clustering individuals into a small number of discrete fair representations. Discrete representations, however, lack the representational power of distributed representations, which vendors desire. In order to learn distributed fair representations, \citet{edwards2015censoring}, \citet{eissman2018bayesian} and \citet{madras2018learning} each propose adversarial training, where the latter (LAFTR) connects different adversarial losses to multiple notions of fairness. \citet{louizos2015the} propose VFAE for learning distributed fair representations by using a variational autoencoder architecture with additional regularization based on Maximum Mean Discrepancy (MMD)~\citep{gretton2007a}. 
Each of these methods is limited to the case of a binary sensitive attribute because their measurements of fairness are based on statistical parity~\citep{zemel2013learning}, which is defined only for two groups.

Interestingly, each of these methods can  be viewed as optimizing an \textit{approximation} of the Lagrangian dual of our objective in Equation~\ref{eq:tractable-objective}, with particular \textit{fixed} settings of the Lagrangian multipliers:
\begin{align}
  & \argmin_{\theta, \phi} \max_\psi \mc{L}_r + \lambda_1 (C_1-\epsilon_1) + \lambda_2 (C_2-\epsilon_2) \label{eq:mifr} \\
=  & \argmin_{\theta, \phi} \max_\psi \mc{L}_r + \lambda_1 C_1 + \lambda_2 C_2 \nonumber
\end{align}
where $\mc{L}_r$, $C_i$ and $\epsilon_i$ are defined as in Equation~\ref{eq:tractable-objective}, 
and the multipliers $\lambda_i \geq 0$ are hyperparameters controlling the relative strengths of the constraints (which now act as ``soft'' regularizers). 

We use ``approximation'' to suggest these objectives are not exactly the same as ours, as ours can deal with more than two groups in the fairness criterion $C_2$ and theirs cannot. However, all the fairness criteria achieve $\zv \perp \uv$ at a global optimum; in the following discussions, for brevity we use $C_2$ to indicate their objectives, even when they are not identical to ours\footnote{We also have not included the task classification error in their methods, as we do not assume a single, specific task or assume access to labels in our setting.}.

Here, the values of $\epsilon$ 
do not affect the final solution. 
Therefore, if we wish to find representations that satisfy specific constraints, we would have to search over the hyperparameter space to find feasible solutions, which could be computationally inefficient. 
We call this class of approaches \textit{\methodlong} (\method\footnote{Pronounced \href{https://zelda.gamepedia.com/Mipha}{``Mipha''}.}). 
In Table~\ref{tab:summary}, we summarize these existing methods. 

\begin{table}[htbp]
    \centering
    \caption{Summarizing the components in existing methods. The hyperparameters (e.g. $A_z$, $\alpha$, $\beta$) are from the original notations of the corresponding methods. 
    }
    \begin{tabular}{|*{3}{c|}}
    \toprule
               & $\lambda_1$ & $\lambda_2$ \\
            

    \midrule
    \cite{zemel2013learning}       &  0 & $A_z / A_x$ \\
    \cite{edwards2015censoring}    &  0 & $\alpha / \beta$ \\
    \cite{madras2018learning}      &  0 & $\gamma / \beta$ \\
    \cite{louizos2015the}          &  1 & $\beta$ \\
    \bottomrule
    \end{tabular}
    \label{tab:summary}
\end{table}




\begin{itemize}
\item \citet{zemel2013learning} consider $\mc{L}_r$ as well as minimizing statistical parity (Equation 4 in their paper); they assume $\zv$ is discrete, bypassing the need for adversarial training. Their objective is equivalent to Equation~\ref{eq:mifr} with $\lambda_1 = 0, \lambda_2 = A_z / A_x$. 
\item \citet{edwards2015censoring} considers $\mc{L}_r$ (where $\pxizu$ is Gaussian) and adversarial training where the adversary tries to distinguish the representations from two groups (Equation 9). Their objective is equivalent to Equation~\ref{eq:mifr} with $\lambda_1 = 0, \lambda_2 = \alpha / \beta$. 
\item \citet{madras2018learning} considers $\mc{L}_r$ and adversarial training, which optimizes over surrogates to the demographic parity distance between two groups (Equation 4). Their objective is equivalent to Equation~\ref{eq:mifr} with $\lambda_1 = 0, \lambda_2 = \gamma / \beta$. 
\item \citet{louizos2015the} consider $\mc{L}_r$, $C_1$ with $\lambda_1 = 1$ and the maximum mean discrepancy between two sensitive groups ($C_2$) (Equation 8). However, as $\mc{L}_r + C_1$ is the VAE objective, their solutions does not prefer high mutual information between $\xv$ and $\zv$ (referred to as the ``information preference'' property~\citep{chen2016variational,zhao2017towards,zhao2017infovae,zhao2018a}). Their objective is equivalent to Equation~\ref{eq:mifr} with $\lambda_1 = 1, \lambda_2 = \beta$. 
\end{itemize}
All of the above methods requires hand-tuning $\lambda$ to govern the trade-off between the desiderata, 
because each of these approaches optimizes the dual with \textit{fixed} multipliers instead of \textit{optimizing} the multipliers to satisfy the fairness constraints, $\epsilon$ is ignored, so these approaches cannot ensure that the fairness constraints are satisfied. Using any of these approaches to empirically achieve a desirable limit on unfairness requires manually tuning the multipliers (e.g., increase some $\lambda_i$ until the corresponding constraint is satisfied) over many experiments and is additionally difficult because there is no interpretable relationship between the multipliers and a \textit{limit} on unfairness.

Our method is also related to other works on fairness and information theory. \citet{komiyama2018nonconvex} solve least square regression under multiple fairness constraints. \citet{calmon2017optimized} transform the dataset to prevent discrimination on specific classification tasks. \citet{zhao2018a} discussed information-theoretic constraints in the context of learning latent variable generative models, but did not discuss fairness.

\section{DUAL OPTIMIZATION FOR CONTROLLABLE FAIR REPRESENTATIONS}
\label{sec:dual}

In order to exactly solve the dual of our practical objective from Equation~\ref{eq:tractable-objective} and guarantee that the fairness constraints are satisfied, we must optimize the model parameters as well as the Lagrangian multipliers, which we do using the following dual objective:
\begin{align}
   \max_{\lambdav \geq 0} \min_{\theta, \phi} \max_{\psi} \mc{L} = \mc{L}_r + \lambdav^\top (\Cv - \epsilonv) \label{eq:lag-dual}
\end{align}
where $\lambdav = [\lambda_1, \lambda_2]$ are the multipliers and $\epsilonv = [\epsilon_1, \epsilon_2]$ and $\Cv = [C_1, C_2]$ represent the constraints. 

If we assume we are optimizing in the distribution space (i.e. $\Phi,\Theta$ corresponds to the set of all valid distributions $(\qzixu, \pxizu, p_\theta(\zv))$), then we can show that strong duality holds (our primal objective from Equation~\ref{eq:tractable-objective} equals our dual objective from Equation~\ref{eq:lag-dual}). 
\done{\rk{In C2 below, $p_\phi$ is used, doesn't match the $p_\psi$ used when stating the same objective earlier} \js{exactly. this is where we assume the adversary is optimal; i wonder if we can add $p_\psi$ into the $C_2$ and say stuff; maybe}
\rk{Haven't double-checked or rederived or edited this theorem}}
\begin{theorem}\label{thm:slater}
If $\epsilon_1, \epsilon_2 > 0$, then strong duality holds for the following optimization problem over distributions $p_\theta$ and $q_\phi$:
\begin{align}
 \min_{p_\theta, q_\phi} \quad  & - \bb{E}_{\qxzu}[\log \pxizu] \\
\text{s.t.} \quad & \bb{E}_{\qxu}\KL(\qzixu \Vert p_\theta(\zv)) < \epsilon_1 \nonumber \\
    & \bb{E}_{\qz}\KL(\quiz \Vert \pu) < \epsilon_2 \nonumber \\
\end{align}
where $q_\phi$ denotes $\qzixu$ and $p_\theta$ denotes $p_\theta(\zv)$ and $\pxizu$.
\end{theorem}
We show the complete proof in Appendix~\ref{app:slater}. Intuitively, we utilize the convexity of KL divergence (over the pair of distributions) and mutual information (over the conditional distribution) to verify that Slater's conditions hold for this problem. 




In practice, we can perform standard iterative gradient updates in the parameter space: standard gradient descent over $\theta, \phi$, gradient ascent over $\psi$ (which parameterizes only the adversary), and gradient ascent over $\lambdav$.
Intuitively, the gradient ascent over $\lambdav$ corresponds to a multiplier $\lambdav$ increasing when its constraint is not being satisfied, encouraging the representations to satisfy the fairness constraints even at a cost to representation expressiveness. Empirically, we show that this scheme is effective despite non-convexity in the parameter space.

Note that given finite model capacity, an $\epsilonv$ that is too small may correspond to no feasible solutions in the parameter space; that is, it may be impossible for the model to satisfy the specified fairness constraints. Here we introduce heuristics to estimate the mimimum feasible $\epsilonv$. The minimum feasible $\epsilon_1$ and $\epsilon_3$ can be estimated by running the standard conditional VAE algorithm on the same model and estimating the value of each divergence. Feasible $\epsilon_2$ can be approximated by $\hq(\uv)$, since $\iq(\zv; \uv) \leq \hq(\uv)$; This can easily be estimated empirically when $\uv$ is binary or discrete.


\section{EXPERIMENTS}
\label{sec:experiments}

We aim to experimentally answer the following:
\begin{itemize}
    \item Do our information-theoretical objectives align well with existing notions of fairness?
    \item Do our constraints achieve their intended effects?
    \item How do MIFR and L-MIFR compare when learning controllable fair representations?
    \item How are the learned representations affected by other hyperparameters, such as the number of iterations used for adversarial training in $C_2$?
    \item Does L-MIFR have the potential to balance different notions of fairness?
\end{itemize}

\subsection{Experimental Setup}
We evaluate our results on three datasets~\citep{zemel2013learning,louizos2017causal,madras2018learning}. 
The first is the UCI \textit{German} credit dataset\footnote{\href{https://archive.ics.uci.edu/ml/datasets}{https://archive.ics.uci.edu/ml/datasets}}, which contains information about 1000 individuals, with a binary sensitive feature being whether the individual's age exceeds a threshold. The downstream task is to predict whether the individual is offered credit or not.
The second is the UCI \textit{Adult} dataset\footnote{\href{https://archive.ics.uci.edu/ml/datasets/adult}{https://archive.ics.uci.edu/ml/datasets/adult}}, which contains information of over 40,000 adults from the 1994 US Census. The downstream task is to predict whether an individual earns more than \$50K/year. We consider the sensitive attribute to be gender, which is pre-processed to be a binary value. The third is the Heritage \textit{Health} dataset\footnote{\href{https://www.kaggle.com/c/hhp}{https://www.kaggle.com/c/hhp}}, which contains information of over 60,000 patients. The downstream task is to predict whether the Charlson Index (an estimation of patient mortality) is greater than zero. Diverging from previous work~\citep{madras2018learning}, we consider sensitive attributes to be age and gender, where there are 9 possible age values and 2 possible gender values; hence the sensitive attributes have 18 configurations. This prevents VFAE~\citep{louizos2015the} and LAFTR~\citep{madras2018learning} from being applied, as both methods reply on some statistical distance between two groups, which is not defined when there are 18 groups in question\footnote{$\Delta_{DP}$ is only defined for binary sensitive variables in~\citep{madras2018learning}.}. 

We assume that the model does not have access to labels during training; instead, it supplies its representations to an unknown vendor's classifier, whose task is to achieve high prediction with labels. We compare the performance of \emph{\method}, the model with fixed multipliers, and \emph{L-\method}, the model using the Lagrangian dual optimization method. We provide details of the experimental setup in Appendix~\ref{sec:exp}. Specifically, we consider the simpler form for $\pz$ commonly used in VAEs, where $\pz$ is a fixed prior; the use of other more flexible parametrized forms of $\pz$, such as normalizing flows~\citep{dinh2016density,rezende2015variational} and autoregressive models~\citep{kingma2016improving,oord2016pixel}, is left as future work.


We estimate the mutual information values $\iq(\xv; \zv |\uv)$ and $\iq(\uv; \zv)$ on the test set using the following equations:
\begin{gather*}
\iq(\xv; \zv |\uv) = \bb{E}_{\qxzu}[\log \qzixu - \log \qziu] \\
\iq(\uv; \zv) = \bb{E}_{\qxzu}[\log q_\phi(\uv | \zv) - \log q(\uv)]
\end{gather*}
where $\qziu$ is estimated via kernel density estimation over samples from $\qzixu$ with $(\xv, \uv)$ sampled from the training set. Kernel density estimates are reasonable since both $\zv$ and $\uv$ are low dimensional (for example, \textit{Adult} considers a 10-dimension $\zv$ for 40,000 individuals). 
However, computing $\qziu$ requires a summation over the training set, so we only compute these mutual information quantities during evaluation. We include our implementations in \href{https://github.com/jiamings/lag-fairness}{https://github.com/ermongroup/lag-fairness}.

\subsection{Mutual Information, Prediction Accuracy, and Fairness}
\begin{figure}[htbp]
    \centering
    \begin{subfigure}{0.40\textwidth}
    \includegraphics[width=\textwidth]{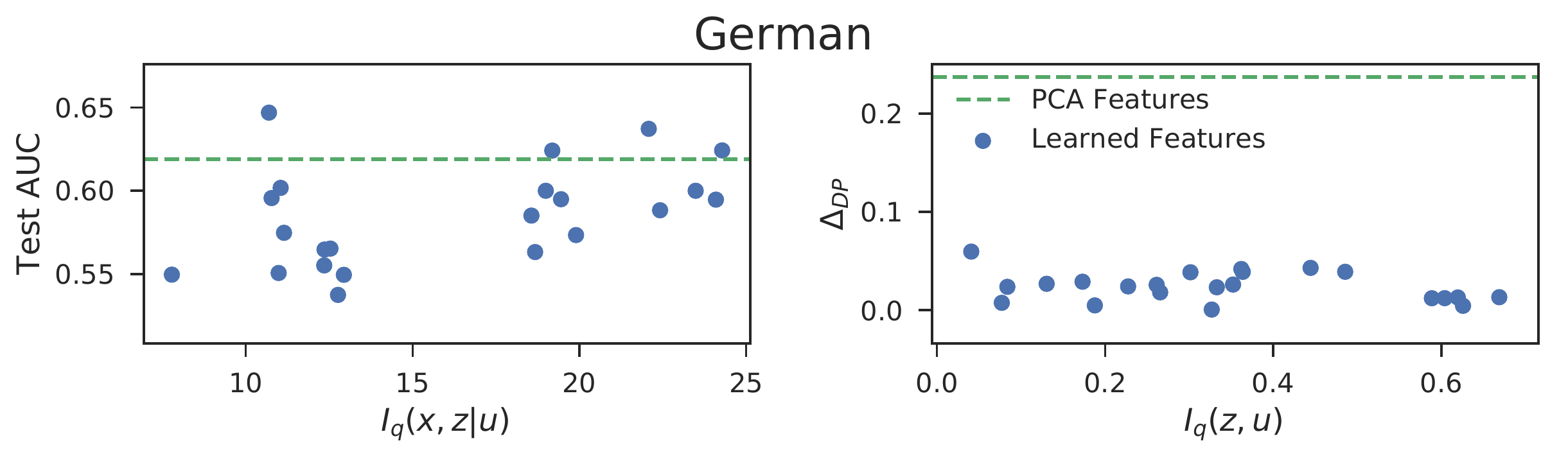}
    \end{subfigure}
    ~
    \begin{subfigure}{0.40\textwidth}
    \includegraphics[width=\textwidth]{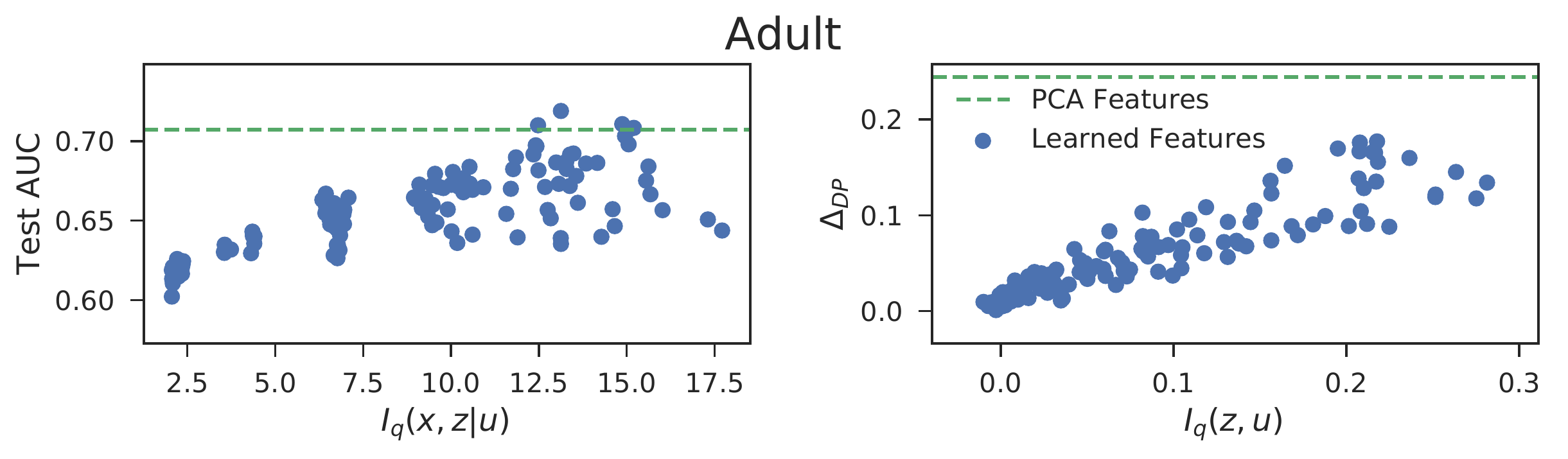}
    \end{subfigure}
    ~
    \begin{subfigure}{0.40\textwidth}
    \includegraphics[width=0.5\textwidth]{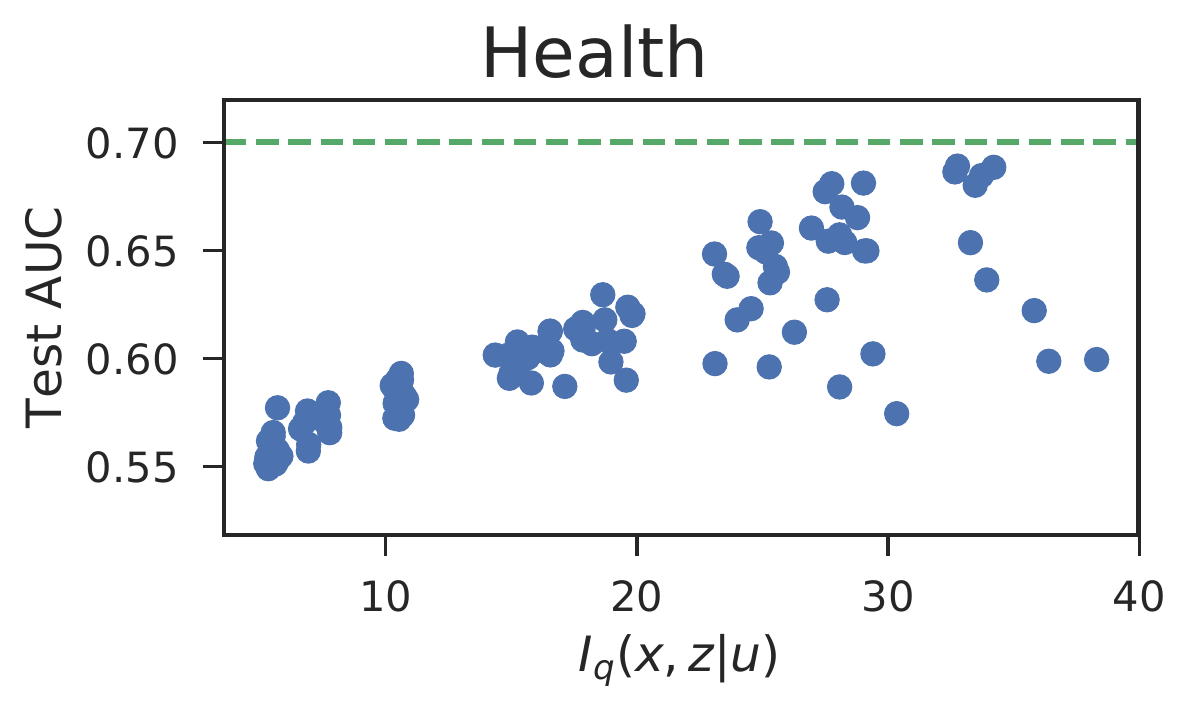}
    \end{subfigure}
    \caption{The relationship between mutual information and fairness related quantities. Each dot is the representations from an instance of MIFR with a different set of hyperparameters. Green line represents features obtained via principle component analysis. Increased mutual information between inputs and representations increase task performance (left) and unfairness (right). For \textit{Health} we do not include $\Delta_{DP}$ since it is not defined for more than two groups.
    }
    \label{fig:mifr_hypers}
\end{figure}
\begin{figure}[htbp]
    \centering
    \begin{subfigure}{0.40\textwidth}
    \includegraphics[width=\textwidth]{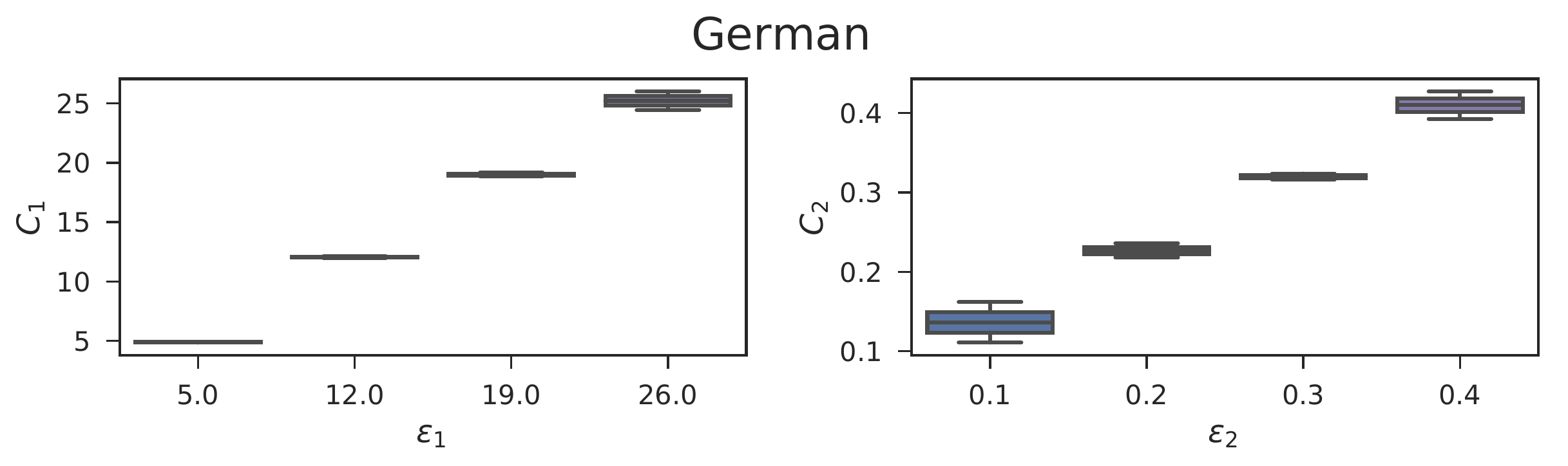}
    \end{subfigure}
    ~
    \begin{subfigure}{0.40\textwidth}
    \includegraphics[width=\textwidth]{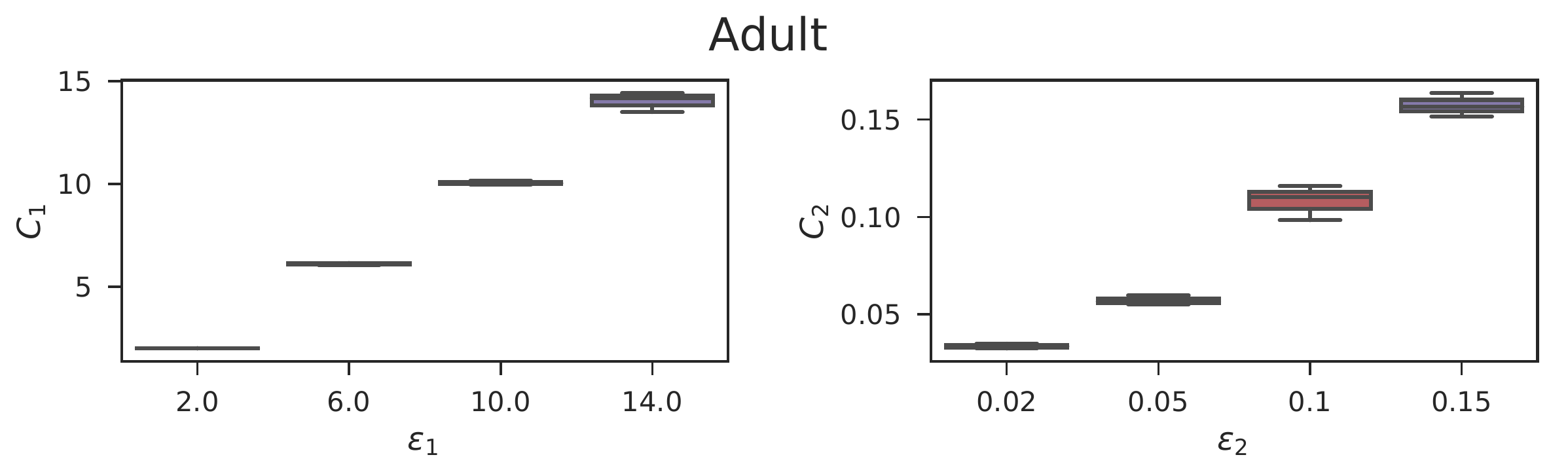}
    \end{subfigure}
    ~
    \begin{subfigure}{0.40\textwidth}
    \includegraphics[width=\textwidth]{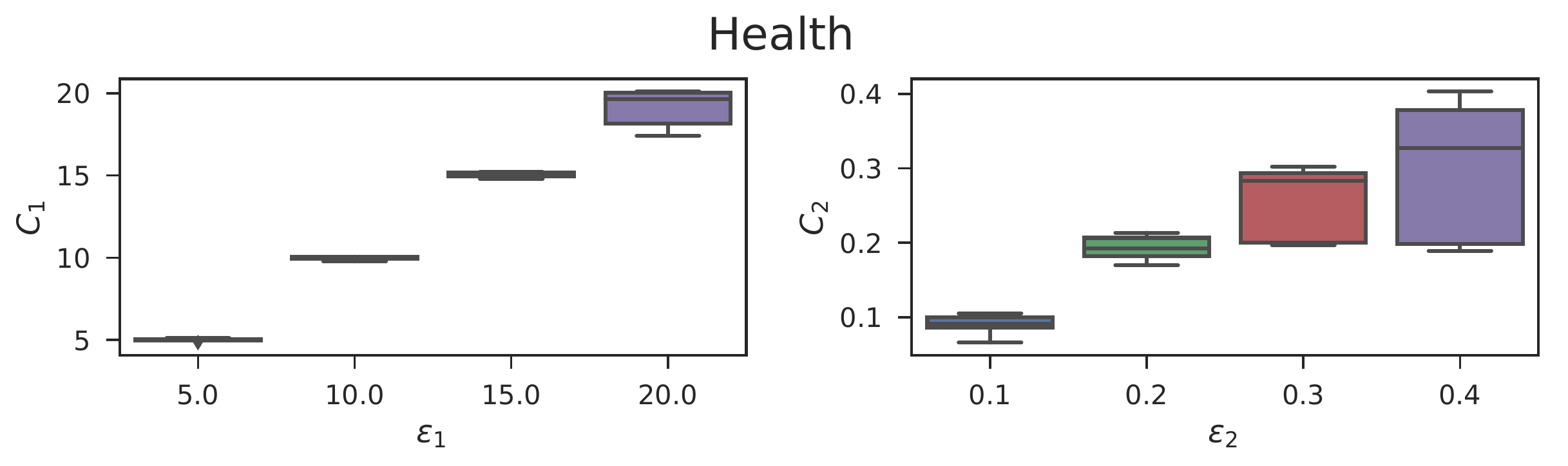}
    \end{subfigure}
    \caption{Corresponding $C_i$ values under different $\epsilon_i$ with L-\method. After $\epsilon_i$ is fixed, we consider a range of values for the other constraint, leading to a distribution of $C_i$ for each $\epsilon_i$ (hence the box plot). 
    }
    
    \label{fig:lmifr_hypers}
\end{figure}



We investigate the relationship between mutual information and prediction performance by considering 
area under the ROC curve (AUC) for prediction tasks. We also investigate the relationship between mutual information and traditional fairness metrics by considering the $\Delta_{DP}$ fairness metric in~\cite{madras2018learning}, which compares the absolute expected difference in classifier outcomes between two groups. $\Delta_{DP}$ is only defined on two groups of classifier outcomes, so it is not defined for the \textit{Health} dataset when considering the sensitive attributes to be ``age and gender'', which has 18 groups.
We use logistic regression classifiers for prediction tasks. 


From the results results in Figure~\ref{fig:mifr_hypers}, we show that there are strong positive correlations 
between $\iq(\xv; \zv | \uv)$ and test AUC, and between $\iq(\zv, \uv)$ and $\Delta_{DP}$; increases in $\iq(\zv, \uv)$ decrease fairness. We also include a baseline in Figure~\ref{fig:mifr_hypers} where the features are obtained via the top-$k$ principal components (where $k$ is the dimension of $\zv$), which has slightly better AUC but significantly worse fairness as measured by $\Delta_{DP}$. 
Therefore, our information theoretic notions of fairness/expressiveness align well with existing notions such as $\Delta_{DP}$/test AUC. 

\subsection{Controlling Representation Fairness with L-\method}



 Keeping all other constraint budgets fixed, any increase in $\epsilon_i$ for an arbitrary constraint $C_i$ implies an increase in the unfairness budget; consequently, we are able to trade-off fairness for more informative representations when desired.

We demonstrate this empirically via an experiment where we note the $C_i$ values corresponding to a range of budgets $\epsilon_i$ at a fixed configuration of the other constraint budgets $\epsilon_j$ ($j \neq i$). 
From Figure~\ref{fig:lmifr_hypers}, $C_i$ increases as $\epsilon_i$ increases, and $C_i < \epsilon_i$ holds under different values of the other constraints $\epsilon_j$. This suggest that we can use $\epsilon_i$ to control $C_i$ (our fairness criteria) of the learned representations. 

\begin{figure}
    \centering
    \begin{subfigure}{0.40\textwidth}
    \includegraphics[width=\textwidth]{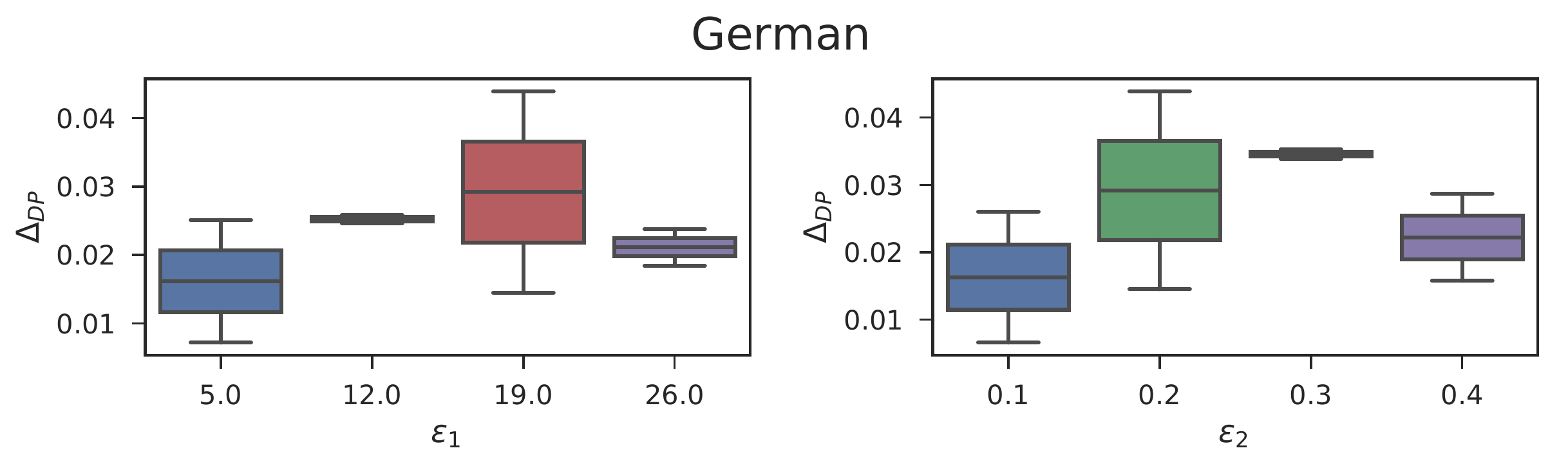}
    \end{subfigure}
    ~
    \begin{subfigure}{0.40\textwidth}
    \includegraphics[width=\textwidth]{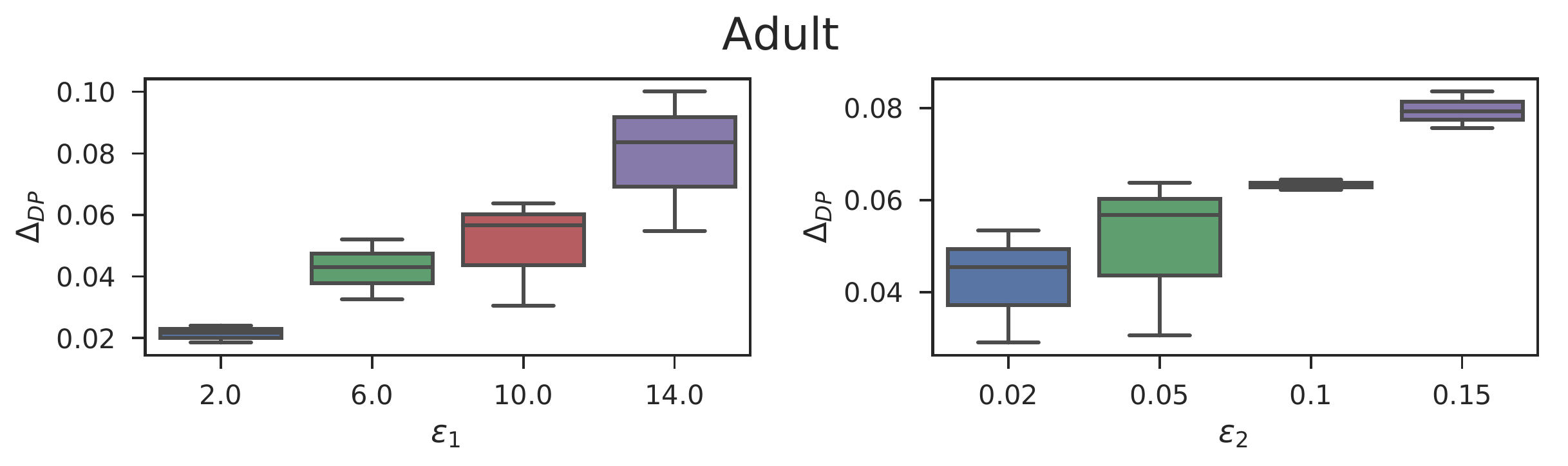}
    \end{subfigure}
    \caption{$\Delta_{DP}$ under different levels of $\epsilon$ with L-MIFR. $\Delta_{DP}$ generally increases as $\epsilon$ increases.}
    
    \label{fig:lmifr_dp}
\end{figure}

We further show the changes in $\Delta_{DP}$ (a traditional fairness criteria) values as we vary $\epsilon_i$ in Figure~\ref{fig:lmifr_dp}. In \textit{Adult}, $\Delta_{DP}$ clearly increases as $\epsilon_i$ increases; this is less obvious in \textit{German}, as $\Delta_{DP}$ is already very low. These results suggest that the L-MIFR user can control the level of fairness of the representations quantitatively via $\epsilon$. 






\subsection{Improving Representation Expressiveness with L-\method}

Recall that our goal is to perform controlled fair representation learning, which requires us to learn expressive representations subject to fairness constraints. We compare two approaches that could achieve this: 1) MIFR, which has to consider a range of Lagrange multipliers (e.g. from a grid search) to obtain solutions that satisfy the constraints; 2) L-MIFR, which finds feasible solutions directly by optimizing the Lagrange multipliers.

\begin{figure}[htbp]
    \centering
    \includegraphics[width=0.43\textwidth]{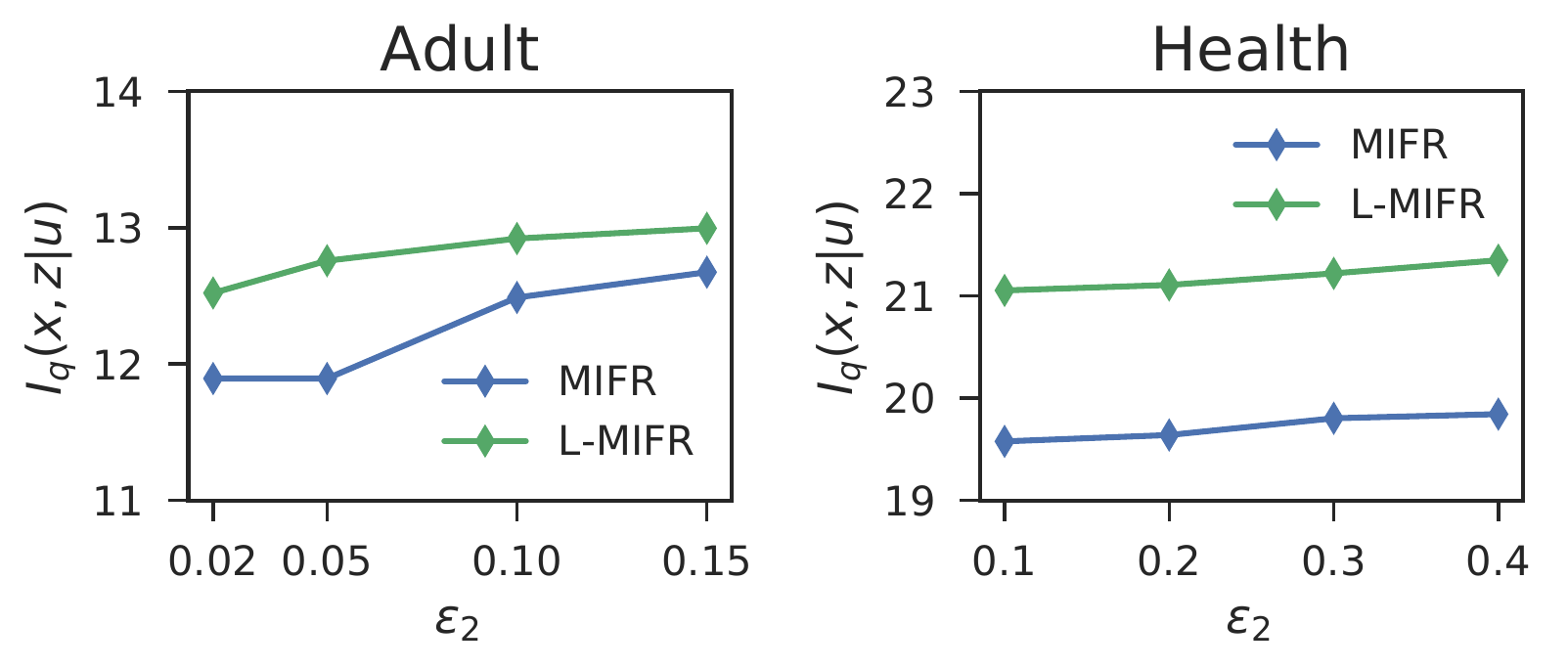}
    \caption{Expressiveness vs. $\epsilon_2$. A larger feasible region (as measured by $\epsilon_2$) leads to more expressive representations (as measured by $\iq(\xv, \zv|\uv)$). 
    }
    \label{fig:expressive}
\end{figure}

We evaluate both methods on 4 sets of constraints by modifying the values of $\epsilon_2$ (which is the tighter estimate of $\iq(\zv; \uv)$) while keeping $\epsilon_1$ fixed, and we compare the expressiveness of the features learned by the two methods in Figure~\ref{fig:expressive}. For MIFR, we perform a grid search running $5^2 = 25$ configurations. 
In contrast, we run \textit{one} instance of L-MIFR for each $\epsilon$ setting, which takes roughly the same time to run as one instance of MIFR (the only overhead is updating the two scalar values $\lambda_1$ and $\lambda_2$). 

In terms of representation expressiveness, L-MIFR outperforms MIFR even though MIFR took almost 25x the computational resources. 
Therefore, L-MIFR is significantly more computationally efficient than MIFR at learning controlled fair representation. 

\subsection{Ablation Studies}
The $C_2$ objective requires adversarial training, which involves iterative training of $(\theta, \phi)$ with $\psi$. We assess the sensitivity of the expressiveness and fairness of the learned representations to the number of iterations $D$ for $\psi$ per iteration for $(\theta, \phi)$. Following practices in~\citep{gulrajani2017improved} to have more iterations for critic, we consider $D = \{1, 2, 5, 10\}$, and use the same number of total iterations for training.

\begin{table}
    \centering
    
    \begin{tabular}{c|c|cccc}
    \toprule

        & & $D=$1 & $D=$2 & $D=$5 & $D=$10  \\
    \midrule
    \multirow{2}{*}{Adult}  & $\iq(\xv; \zv|\uv)$ & $10.46$ & $10.94$ & $9.75$ & $9.54$ \\
    & $\iq(\zv; \uv)$ & $0.10$ & $0.07$ & $0.08$ & $0.06$ \\\midrule
    \multirow{2}{*}{Health}  & $\iq(\xv; \zv|\uv)$ & $16.60$ & $16.47$ & $16.65$ & $16.75$ \\
    & $\iq(\zv; \uv)$ & $0.17$ & $0.17$ & $0.22$ & $0.28$ \\\bottomrule
    \end{tabular}
    \caption{Expressiveness and fairness of the representations from L-MIFR under various $D$.}
    \label{tab:adversarial}
\end{table}

In Table~\ref{tab:adversarial}, we evaluate $\iq(\xv; \zv|\uv)$ and $\iq(\zv; \uv)$ obtained L-MIFR on \textit{Adult} ($\epsilon_2 = 0.10$) and \textit{Health} ($\epsilon_2 = 0.30$). This suggests that the final solution of the representations is not very sensitive to $D$, although larger $D$ seem to find solutions that are closer to $\epsilon_2$. 

\subsection{Fair Representations under Multiple Notions}
\label{sec:exp:multi}

Finally, we demonstrate how L-MIFR could control multiple fairness constraints simultaneously, thereby finding representations that are reasonably fair when there are multiple fairness notions being considered. We consider the \textit{Adult} dataset, and describe the \textit{demographic parity}, \textit{equalized odds} and \textit{equalized opportunity} notions of fairness in terms of mutual information, which we denote as $I_{DP} := \iq(\zv; \uv)$, $I_{EO}$, $I_{EOpp}$ respectively (see details in Appendix~\ref{app:eodds-eopp} about how $I_{EO}$ and $I_{EOpp}$ are derived). 

For L-MIFR, we set $\epsilon_1 = 10$ and other $\epsilon$ values to $0.1$. For MIFR, we consider a more efficient approach than random grid search. We start by setting every $\lambda = 0.1$; then we multiply the $\lambda$ value for a particular constraint by $2$ until the constraint is satisfied by MIFR; we finish when all the constraints are satisfied\footnote{This allows MIFR to approach the feasible set from outside, so the solution it finds will generally have high expressiveness.}. We find that this requires us to update the $\lambda$ of $I_{DP}$, $I_{EO}$ and $I_{EOpp}$ four times each (so corresponding $\lambda = 1.6$); this costs 12x the computational resources needed by L-MIFR.

\begin{table}
    \centering
    \begin{tabular}{c|ccccc}
    \toprule
       & $\iq(\xv; \zv|\uv)$ & $C_1$ & $I_{DP}$ &  $I_{EO}$ & $I_{EOpp}$ \\\midrule
       MIFR & 9.34 & \textbf{9.39} &  0.09 & 0.10 & 0.07 \\
       L-MIFR & \textbf{9.94} & 9.95 & \textbf{0.08} & \textbf{0.09} & \textbf{0.04} \\
    \bottomrule
    \end{tabular}
    \caption{Learning one representation for multiple notions of fairness on \textit{Adult}. L-MIFR learns representations that are better than MIFR on all the measurements instead of only $C_1$. Here $\epsilon_1 = 10$ for $C_1$ and $\epsilon = 0.1$ for other constraints.}
    \label{tab:notions}
\end{table}

We compare the representations learned by L-MIFR and MIFR in Figure~\ref{tab:notions}. L-MIFR outperforms MIFR in terms of $\iq(\xv; \zv|\uv)$, $I_{DP}$, $I_{EO}$ and $I_{EOpp}$, while only being slightly worse in terms of $C_1$. Since $\epsilon_1 = 10$, the L-MIFR solution is still feasible. This demonstrates that even with a thoughtfully designed method for tuning $\lambda$, MIFR is still much inferior to L-MIFR in terms of computational cost and representation expressiveness. 

\section{DISCUSSION}

In this paper, we introduced an objective for learning controllable fair representations based on mutual information. This interpretation allows us to unify and explain existing work. In particular, we have shown that a range of existing approaches optimize an approximation to the Lagrangian dual of our objective with \textit{fixed} multipliers, fixing the trade-off between fairness and expressiveness. 
We proposed a dual optimization method 
that allows us to achieve higher expressiveness while satisfying the user-specified limit on unfairness.


In future work, we are interested in formally and empirically extending this framework and the corresponding dual optimization method to other notions of fairness. 
It is also valuable to investigate alternative approaches to training the adversary~\citep{gulrajani2017improved}, the usage of more flexible $p(\zv)$~\citep{rezende2015variational}, and alternative solutions to bounding $\iq(\zv, \uv)$.

\newpage
\subsubsection*{Acknowledgements}
This research was supported by NSF (\#1651565, \#1522054, \#1733686), ONR (N00014-19-1-2145), AFOSR (FA9550-19-1-0024), and FLI. We thank Xudong Shen for helpful discussions.

\bibliography{ref}
\bibliographystyle{plainnat}

\newpage
\clearpage
\onecolumn
\appendix
\section{Proofs}\label{sec:proofs}
\subsection{Proof of Lemma 1}
\begin{proof}
\begin{align*}
\iq(\xv; \zv | \uv) & = \bb{E}_{\qxzu}[\log \qxziu - \log \qxiu - \log \qziu] \\
& = \bb{E}_{\qxzu}[\log \qxziu  - \log \qziu] + \bb{E}_{\qzu}[- \log \qxiu] \\
& = \bb{E}_{\qxzu}[\log \qxizu] + \hq(\xv | \uv) \\
& = \bb{E}_{\qxzu}[\log \qxizu + \log p(\xv|\zv, \uv) - \log p(\xv|\zv, \uv)] + \hq(\xv | \uv) \\
& = \bb{E}_{\qxzu}[\log p(\xv|\zv, \uv)] + \hq(\xv | \uv) + \bb{E}_{\qzu}\KL(\qxizu \Vert p(\xv|\zv, \uv)) \\
& \geq \bb{E}_{\qxzu}[\log p(\xv|\zv, \uv)] + \hq(\xv | \uv)
\end{align*}
\done{\rk{Do we need to define $D_{KL}$ once?} \js{fixed}}
where the last inequality holds because KL divergence is non-negative.
\end{proof}

\subsection{Proof of Lemma 2}
\begin{proof}
\begin{align*}
    \iq(\zv; \uv) & \leq \iq(\zv; \xv, \uv) \\
& = \bb{E}_{\qxzu}[\log \qzixu - \log \qz] \\
& = \bb{E}_{\qxzu}[\log \qzixu - \log \pz - \log \qz + \log \pz] \\
& = \bb{E}_{\qxu}\KL(\qzixu \Vert \pz) - \KL(\qz \Vert \pz)
\end{align*}
\end{proof}

\subsection{Proof of Lemma 3}
\label{app:c2}
\begin{proof}
\begin{align*}
    \iq(\zv; \uv) & = \bb{E}_{\qzu}[\log \quiz - \log \qu] \\
    & = \bb{E}_{\qzu}[\log \quiz - \log \pu - \log \qu + \log \pu] \\
    & = \bb{E}_{\qz}\KL(\quiz \Vert \pu) - \KL(\qu \Vert \pu) \\
    & \leq \bb{E}_{\qz}\KL(\quiz \Vert \pu)
\end{align*}
Again, the last inequality holds because KL divergence is non-negative.
\done{\rk{We need to introduce $\quiz$?}\js{fixed}}
\end{proof}


\subsection{Proof of Theorem~\ref{thm:slater}} \label{app:slater}
\begin{proof}
Let us first verify that this problem is convex.
\begin{itemize}
    \item Primal: $-\bb{E}_{\qxzu}[\log \pxizu]$ is affine in $\qzixu$, convex in $\pxizu$ due to the concavity of $\log$, and independent of $p_\theta(\zv)$.
    \item First condition: $\bb{E}_{\qu}\KL(\qzixu \Vert p_\theta(\zv))$ is convex in $\qzixu$ and $p_\theta(\zv)$ (because of convexity of KL-divergence), and independent of $\pxizu$.
    \item Second condition: since $\bb{E}_{\qz}\KL(\quiz \Vert \pu) - \KL(\qu \Vert \pu) = \iq(\zv; \uv)$ and
    \begin{align}
        \iq(\zv; \uv) & = \KL(\qzu \Vert \qu \qz) \\
        & = \KL( \sum_\xv \qzixu \qxu \Vert \qu \sum_{\xv, \uv} \qzixu \qxu )
    \end{align}
    Let $q = \beta q_1 + (1 - \beta) q_2$, $\forall \beta \in [0, 1], q_1, q_2$. We have
    \begin{align*}
        I_{q}(\zv; \uv) &= \KL( \sum_\xv q(\zv|\xv, \uv) \qxu \Vert \qu \sum_{\xv, \uv} q(\zv|\xv, \uv) \qxu ) \\
        & \geq \beta \KL( \sum_\xv q_1(\zv|\xv, \uv) \qxu \Vert \qu \sum_{\xv, \uv} q_1(\zv|\xv, \uv) \qxu ) \\
        & \qquad + (1 - \beta) \KL( \sum_\xv q_2(\zv|\xv, \uv) \qxu \Vert \qu \sum_{\xv, \uv} q_2(\zv|\xv, \uv) \qxu ) \\
        & = \beta I_{q_1}(\zv; \uv) + (1 - \beta) I_{q_2}(\zv; \uv)
    \end{align*}
    where we use the convexity of KL divergence in the inequality. Since $\KL(\qu \Vert \pu)$ is independent of $\qzixu$, both $\iq(\zv; \uv)$ and $\bb{E}_{\qz}\KL(\quiz \Vert \pu)$ are convex in $\qzixu$.
\end{itemize}
Then we show that the problem has a feasible solution by construction. 
In fact, we can simply let $\qzixu = p_\theta(\zv)$ be some fixed distribution over $\zv$, and $\pxizu = \qxizu$ for all $\xv, \uv$. In this case, $\zv$ and $\uv$ are independent, so $\KL(\qzixu \Vert p_\theta(\zv)) = 0 < \epsilon_1$, $\KL(\quiz \Vert \pu) = 0 < \epsilon_2$. This corresponds to the case where $\zv$ is simply random noise that does not capture anything in $\uv$.

Hence, Slater's condition holds, which is a sufficient condition for strong duality.
\end{proof}

\section{Experimental Setup Details} \label{sec:exp}
We consider the following setup for our experiments. 
\begin{itemize}
\item For \method, we modify the weight for reconstruction error $\alpha = 1$, as well as $\lambda_1 \in \{0.0, 0.1, 0.2, 1.0, 2.0\}$ and $\lambda_2 \in \{0.1, 0.2, 1.0, 2.0, 5.0\}$ for the constraints, which creates a total of $5^2 = 25$ configurations; $\lambda_1$ values smaller since high values of $\lambda_1$ prefers solutions with low $\iq(\xv; \zv | \uv)$. 

\item For L-\method, we modify $\epsilon_1$ and $\epsilon_2$ according to the estimated values for each dataset. This allows us to claim results that holds for a certain hyperparameter in general (even as other hyperparameter change). 

\item We use the Adam optimizer with initial learning rate $1e-3$ and $\beta_1 = 0.5$ where the learning rate is multiplied by $0.98$ every $1000$ optimization iterations, following common settings for adversarial training~\citep{gulrajani2017improved}.

\item For L-\method, we initialize the $\lambda_i$ parameters to $1.0$, and allow for a range of $(0.01, 100)$. 

\item Unless otherwise specified, we update $\puiz$ $10$ times per update of $\qzixu$ and $\pxizu$. 

\item For \textit{Adult} and \textit{Health} we optimize for 2000 epochs; for \textit{German} we optimize for 10000 epochs (since there are only 1000 low dimensional data points).

\item For both cases, we consider $q_\phi(\zv | \xv, \uv), p_\theta(\xv | \zv, \uv), p_\psi(\uv|\zv)$ as a two layer neural networks with a hidden layer of 50 neurons with softplus activations, and use $\zv$ of dimension 10 for \textit{German} and \textit{Adult}, and 30 for \textit{Health}. For the joint of two variables (i.e. $(\xv, \uv)$) we simply concatenate them at the input layer.
We find that our conclusions are insensitive to a reasonable change in architectures (e.g. reduce number of neurons to 50 and $\zv$ to 25 dimensions).
\end{itemize}

\section{Comparison with LAFTR}

Our work have several notable differences from prior methods (such as LAFTR~\citep{madras2018learning}) that make it hard to compare them directly.
First, we do not assume access to the prediction task while learning the representation, thus our method does not directly include the “classification error” objective. Second, our method is able to deal with any type of sensitive attributes, as opposed to binary ones. 

Nevertheless, we compare the performance of MIFR and LAFTR (Madras et al.) with the demographic parity notion of fairness (measured by $Delta_{DP}$, lower is better). To make a fair comparison, we add a classification error to MIFR during training. MIFR achieves an accuracy of 0.829 and $\Delta_{DP}$ of 0.037, whereas LAFTR achieves an accuracy of 0.821 and $\Delta_{DP}$ of 0.029. This shows that MIFR and LAFTR are comparable in terms of the accuracy / fairness trade-off. MIFR is still useful for sensitive attributes that are not binary, such as Health, which LAFTR cannot handle. 

We further show a comparison of $\Delta_{DP}$, $\Delta_{EO}$, $\Delta_{EOpp}$ between L-MIFR and LAFTR~\citep{madras2018learning}
on the Adult dataset in Table~\ref{tab:laftr}, where L-MIFR is trained with the procedure in Section~\ref{sec:exp:multi}. While LAFTR achieves better fairness on each notion if it is specifically trained for that notion, it often achieves worse performance on other notions of fairness. We note that L-MIFR uses a logistic regression classifier, whereas LAFTR uses a one layer MLP. Moreover, these measurements are also task-specific as opposed to mutual information criterions.

\begin{table}[htbp]
    \centering
    \begin{tabular}{c|ccc}
    \toprule
         &  $\Delta_{DP}$ & $\Delta_{EO}$ & $\Delta_{EOpp}$ \\\midrule
        L-MIFR & 0.057 & 0.123 & 0.026 \\
        LAFTR-DP & 0.029 & 0.244 & 0.027 \\
        LAFTR-EO & 0.125 & 0.074 & 0.037 \\
        LAFTR-EOpp & 0.098 & 0.154 & 0.022 \\\bottomrule
    \end{tabular}
    \caption{Comparison between L-MIFR and LAFTR on $\Delta_{DP}$, $\Delta_{EO}$, $\Delta_{EOpp}$ metrics from \citep{madras2018learning}. While LAFTR achieves better fairness on individual notions if it is trained for that notion, it often trades that with other notions of fairness.}
    \label{tab:laftr}
\end{table}


\section{Extension to Equalized Odds and Equalized Opportunity}\label{app:eodds-eopp}
If we are also provided labels $y$ for a particular task, in the form of $\mc{D}_l = \{\xuyi\}_{i=1}^{M}$, we can also use the representations to predict $y$, which leads to a third condition:
\begin{enumerate}
\setcounter{enumi}{2}
    \item \textbf{Classification} 
    $\zv$ can be used to classify $y$ with high accuracy. 
\end{enumerate}
We can either add this condition to the primal objective in Equation~\ref{eq:constraint-objective}, or add an additional constraint that we wish to have accuracy that is no less than a certain threshold. 

With access to binary labels, we can also consider information-theoretic approaches to \textit{equalized odds} and \textit{equalized opportunity}~\citep{hardt2016equality}. Recall that \textit{equalized odds} requires that the predictor and sensitive attribute are independent conditioned on the label, whereas \textit{equalized opportunity} requires that the predictor and sensitive attribute are independent conditioned on the label being positive. In the case of learning representations for downstream tasks, our notions should consider any classifier over $\zv$.

For \textit{equalized odds}, we require that $z$ and $u$ have low mutual information conditioned on the label, which is $I_q(\zv, \uv | y)$. For \textit{equalized opportunity}, we require that $z$ and $u$ have low mutual information conditioned on the label $y=1$, which is $I_q(\zv, \uv)|_{y=1}$.

We can still apply the upper bounds similar to the case in $C_2$.
For \textit{equalized opportunity} we have
\begin{align*}
\iq(\zv; \uv)|_{y=1} & \leq \bb{E}_{q_\phi(\zv, \uv, y | y = 1)}[\KL(\quizy \Vert \pu)] - \KL(\qu \Vert \pu) := I_{EO}\\
& \leq \bb{E}_{q_\phi(\zv, \uv, y | y = 1)}[\KL(\quizy \Vert \pu)]
\end{align*}
For \textit{equalized odds} we have
\begin{align*}
\iq(\zv; \uv | y) & = q(1) \iq(\zv; \uv)|_{y=1}  + q(0) \iq(\zv; \uv)|_{y=0} := I_{EOpp} \\
& \leq q(1) \bb{E}_{q_\phi(\zv, \uv, y | y = 1)}[\KL(\quizy \Vert \pu)] + q(0) \bb{E}_{q_\phi(\zv, \uv, y | y = 0)}[\KL(\quizy \Vert \pu)]
\end{align*}
which can be implemented by using a separate classifier for each $y$ or using $y$ as input. If $y$ is an input to the classifier, our mutual information formulation of \textit{equalized odds} does not have to be restricted to the case where $y$ is binary.

\end{document}